\documentclass[runningheads]{llncs}

\usepackage{eccv}

\usepackage{xcolor, colortbl}
\usepackage{multirow, multicol}
\usepackage[ruled,vlined,linesnumbered]{algorithm2e}

\usepackage{tabularx}

\usepackage{pifont}

\newcommand{\expnum}[2]{{#1}\mathrm{e}{-#2}}

\newcommand{\beginsupplement}{
    \setcounter{section}{0}
    \renewcommand{\thesection}{\Alph{section}}
    \renewcommand{\theHsection}{\Alph{section}}
    \setcounter{table}{0}
    \renewcommand{\thetable}{S\arabic{table}}
    \setcounter{figure}{0}
    \renewcommand{\thefigure}{S\arabic{figure}}}
    
\definecolor{commentcolor}{RGB}{110,154,155}   
\newcommand{\PyComment}[1]{\ttfamily\footnotesize\textcolor{commentcolor}{\# #1}}  

\newcommand{\PyCode}[1]{\ttfamily\footnotesize\textcolor{black}{#1}} 

\usepackage{eccvabbrv}

\usepackage{graphicx}
\usepackage{booktabs}

\usepackage[accsupp]{axessibility}  

\usepackage[colorlinks]{hyperref}

\usepackage{orcidlink}

\begin{document}

\title{Online Temporal Action Localization with Memory-Augmented Transformer} 

\titlerunning{Online Temporal Action Localization with Memory-Augmented Transformer}

\author{Youngkil Song\inst{*}\inst{}\orcidlink{0009-0001-9565-2102}\qquad
Dongkeun Kim\inst{*}\inst{}\orcidlink{0000-0001-6093-126X}\qquad
Minsu Cho\inst{}\orcidlink{0000-0001-7030-1958}\qquad
Suha Kwak\inst{}\orcidlink{0000-0002-4567-9091}}
\authorrunning{Youngkil Song, Dongkeun Kim, Minsu Cho, and Suha Kwak}

\institute{Pohang University of Science and Technology (POSTECH), South Korea\\
\email{\{songyk,kdk1563,mscho,suha.kwak\}@postech.ac.kr}\\
\url{https://cvlab.postech.ac.kr/research/MATR/}
}

\maketitle
\makeatletter\def\Hy@Warning#1{}\makeatother
\def\thefootnote{*}\footnotetext{Equal contribution.}\def\thefootnote{\arabic{footnote}}
\begin{abstract}
Online temporal action localization (On-TAL) is the task of identifying multiple action instances given a streaming video.
Since existing methods take as input only a video segment of fixed size per iteration, they are limited in considering long-term context and require tuning the segment size carefully. 
To overcome these limitations, we propose memory-augmented transformer (MATR). 
MATR utilizes the memory queue that selectively preserves the past segment features, allowing to leverage long-term context for inference.
We also propose a novel action localization method that observes the current input segment to predict the end time of the ongoing action and accesses the memory queue to estimate the start time of the action. 
Our method outperformed existing methods on two datasets, THUMOS14 and MUSES, surpassing not only TAL methods in the online setting but also some offline TAL methods. 

\end{abstract}
\section{Introduction}

Today, video is one of the most popular types of media, with tons of videos being created and uploaded to online platforms like YouTube and TikTok every second. 
As a result, the demand for video understanding is growing rapidly.
In particular, dealing with untrimmed videos is a key to practical video understanding since most videos in the real world are not trimmed for individual events in advance.
Temporal action localization (TAL) has been extensively studied in this context~\cite{shou2016temporal,gao2017turn,zhao2017temporal,zeng2019graph,liu2020progressive,xu2020g,zhao2021video,zhu2021enriching,xu2021low,lin2021learning,zhang2022actionformer}, with the aim of detecting each action instance in the input untrimmed video by predicting its start time, end time, and action class.

Recently, online TAL (On-TAL), \ie, TAL for streaming video~\cite{kang2021cag}, has gained increasing attention thanks to its potential applications in video surveillance, sports video analysis, and video summarization.
Unlike TAL which demands the entire video as input, On-TAL aims to detect action instances using only input frames observed so far in an online manner. 
Moreover, once action instances are predicted, it is not allowed to modify the results afterward.

\begin{figure}[t]
    \centering
    \includegraphics[width=1\linewidth]{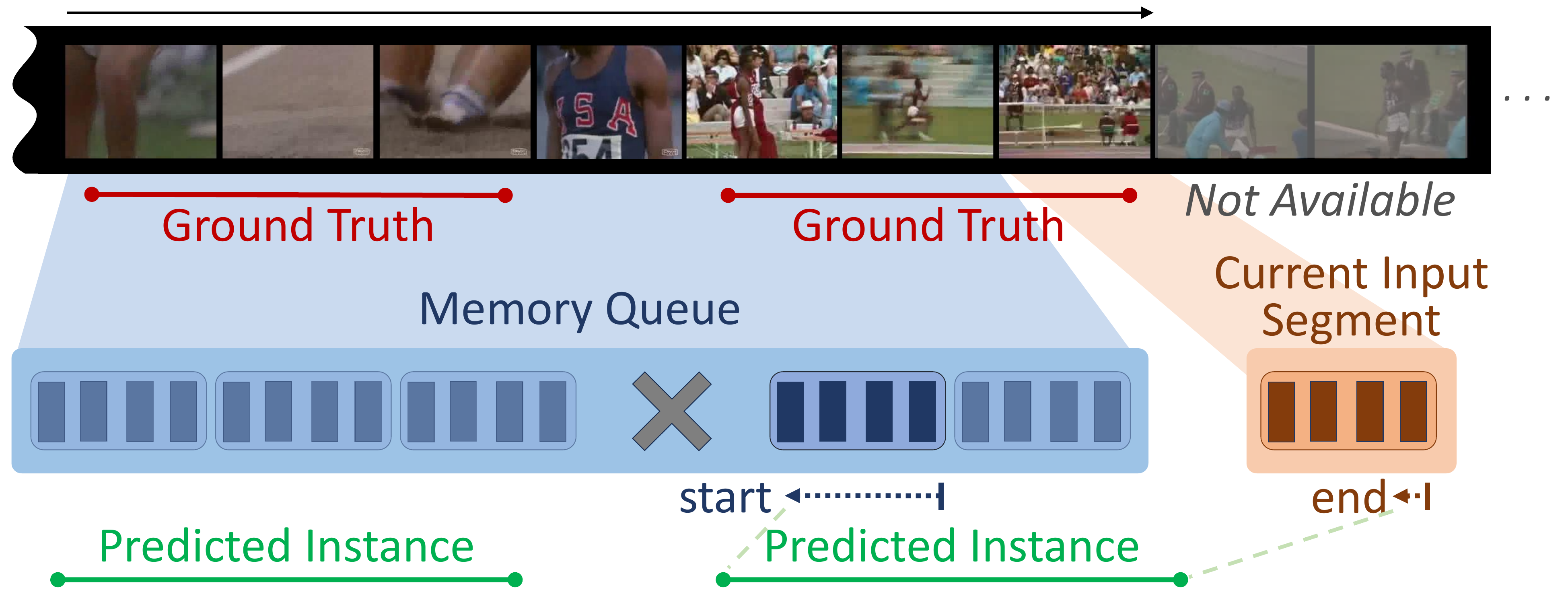}
    \caption{A conceptual illustration of our model. Our model selectively stores information about previously occurred actions in the memory queue from streaming video. When the end of the action is detected in the current input segment, the model retrieves information from the memory queue to locate the start of the action.}
    \label{fig:teaser}
\end{figure}

A straightforward solution to this challenging task is to first conduct online action detection (OAD)~\cite{de2016online,wang2021oadtr}, \ie., frame-by-frame action classification for streaming video,
and then build action instances by aggregating the frame-wise action predictions.
In this sense, earlier methods for On-TAL~\cite{kang2021cag,tang2022simon} follow this approach by applying OAD and refining per-frame predictions using past observations.
However, this OAD-based approach exploits only frame-level supervision for training, which is not optimal because On-TAL aims to predict action instances in the form of time intervals.
To alleviate this issue, online anchor transformer (OAT) has been recently proposed~\cite{kim2022sliding}, which adopts sliding window approach and anchor-based method to utilize instance-level supervision.
Although OAT improved performance remarkably, however, it still has several drawbacks:
Since it takes a fixed-size segment of input video at each iteration and does not consider long-term contexts, its capability is limited in detecting long action instances, and its performance is sensitive to hyperparameters like the size of the input segment.

To resolve the above issues, we propose a new end-to-end architecture for On-TAL, dubbed \emph{memory-augmented} Transformer (MATR). 
At the heart of MATR lies in \emph{memory queue}, which
selectively stores past segment features so that the model exploits long-term context for inference.
Thanks to the memory queue that contains long-term context and reduces dependence on the segment size, our model accurately localizes even long-term action instances without the need for careful hyperparameter tuning for each dataset.

Moreover, we propose a novel approach to action instance localization using the memory queue as illustrated in Fig.~\ref{fig:teaser}.
First, it detects the end of an action using the current segment features and then scans past segment features in the memory queue to find the action start corresponding to the detected action end. 
To achieve this, we adopt two Transformer decoders: one for end detection and the other for start detection. 
Specifically, we adopt learnable queries as input to the decoders so that each of them learns to localize action boundaries through the attention mechanism of Transformer~\cite{vaswani2017attention,dosovitskiy2021an}. 
In addition, motivated by recent advances on object detection~\cite{wu2020rethinking,cheng2022tallformer}, MATR decouples action classification and localization to further improve the performance. 
This is realized in MATR by using distinct queries for the two subtasks. 

The proposed method is evaluated on two datasets, THUMOS14~\cite{THUMOS14} and MUSES~\cite{MUSES}. 
Our model achieves the state of the art on both datasets in the online TAL setting.
In addition, surprisingly, it is also comparable to existing offline TAL methods although it does not employ offline non-maximum suppression (NMS).
In summary, the contribution of this paper is three-fold: 

\begin{itemize}
    \item[$\bullet$] We propose MATR, a new end-to-end architecture for On-TAL, leveraging memory queue to utilize long-term context and enabling precise localization of action instances 
    with less dependence on dataset-specific hyperparameter.
    \item[$\bullet$] We introduce a new action instance localization method that first identifies action end using the input segments and scans past information in the memory queue to find the action start.
    Moreover, our method adopts distinct queries to separate information for action classification and localization. 
    \item[$\bullet$] Our method outperformed the existing On-TAL methods on the two benchmarks. 
    Also, extensive ablation studies demonstrate the contribution of each component of the proposed method.
\end{itemize}

\section{Related Work}
\noindent
\textbf{Temporal Action Localization.}
Temporal action localization (TAL)~\cite{shou2016temporal,gao2017turn,zhao2017temporal,zeng2019graph,chen2019relation,liu2020progressive,xu2020g,zhao2021video,zhu2021enriching,xu2021low,lin2021learning,zhang2022actionformer, liu2022end} which identifies action instances in time and classifying their actions, has given much attention on the video research community. 
Earlier attempts employ two-stage approaches that first generate action proposals, then classify action class and refine action boundaries~\cite{buch2017sst,zhao2017temporal,heilbron2016fast,escorcia2016daps,lin2018bsn,lin2019bmn,liu2019multi}. 
Recent studies have primarily dealt with widening the range of video context through convolutional neural networks~\cite{shou2017cdc}, graph neural networks~\cite{zeng2019graph,xu2020g}, Gaussian kernel~\cite{long2019gaussian}, and utilizing global context~\cite{zhu2021enriching}. 
To extend temporal context and enable end-to-end TAL, Transformer~\cite{vaswani2017attention,dosovitskiy2021an} has recently been adopted~\cite{zhang2022actionformer,liu2022end,cheng2022tallformer}. 
These approaches incorporate temporal deformable attention module~\cite{liu2022end} and multi-scale features~\cite{zhang2022actionformer} to improve computational efficiency and performance.

\noindent
\textbf{Online Video Understanding.}
Online action detection (OAD), which classifies action classes in a frame-wise manner, is first proposed by Geest~\etal~\cite{de2016online}.
To incorporate long-term context, recurrent neural network~\cite{xu2019temporal} and Transformer~\cite{wang2021oadtr} are utilized. 
These methods predict actions for future frames and use this information to refine current frame predictions.
Yang et al.~\cite{yang2022colar} further improve the performance by using category exemplars and capturing long-term dependencies within a video segment.
On the other hand, Shou~\etal~\cite{shou2018online} proposes online detection of action start (ODAS), which aims to detect action start as early as possible.
Gao~\etal~\cite{gao2019startnet} further improves ODAS by adopting reinforcement learning.

\noindent
\textbf{Online Temporal Action Localization.}
Towards practical event understanding, online temporal action localization (On-TAL) has been proposed recently. 
Kang \etal~\cite{kang2021cag} first propose On-TAL, which localizes action boundaries and classifies action classes without accessing future frames and offline post-processing. 
Earlier methods~\cite{kang2021cag,tang2022simon} adopt OAD framework to classify per-frame action class and group them into action instances to solve On-TAL.
Specifically, Kang \etal~\cite{kang2021cag} utilize Markov decision process to aware prior contexts to group per-frame predictions.
SimOn~\cite{tang2022simon} is proposed as an end-to-end On-TAL method that leverages past visual context and action probabilities for accurate action prediction at the current frame.
Kim \etal~\cite{kim2022sliding} propose a sliding window scheme and an anchor-based method that adopts Transformer~\cite{vaswani2017attention} for On-TAL.
However, it only uses a fixed input segment which is highly dependent on the action length distribution of each dataset.
Unlike the previous methods, we utilize a memory queue for long-term context, which also reduces the dependence on the input segment size.

\noindent
\textbf{Difference between OAD and On-TAL.}
OAD~\cite{xu2019temporal,xu2021long,wang2021oadtr,yang2022colar,cao2023e2e, wang2023memory, zhao2022real} focus on predicting action classes at the frame-level, which limits the ability to distinguish boundaries of overlapping actions.
On the other hand, On-TAL~\cite{kang2021cag, tang2022simon, kim2022sliding} predicts entire action instances rather than individual frames, enabling it to precisely identify the start and end times of actions, making it effective for applications requiring instance-level action understanding, like sports video analysis.

\noindent
\textbf{Using Memory for Video Understanding.}
Prior work in video understanding employs memory to store contextual information, in the forms of long-term and short-term memory~\cite{wu2022memvit,xu2021long,cao2023e2e,wang2023memory}: short-term memory captures fine details, and long-term memory stores compressed information of past contexts.
Specifically, Wang~\etal~\cite{wang2023memory} divides long-term memory into several groups and compresses each group into a single vector. 
Stream Buffer~\cite{cao2023e2e} gradually reduces temporal dimensions of the memory using a compression module.
However, unlike video classification tasks such as action recognition~\cite{wu2022memvit} and OAD~\cite{xu2021long,cao2023e2e,wang2023memory,zhao2022real}, which requires to store representative information of past frames in the memory for classifying the action of the current frame, TAL needs to preserve the temporal positional information of past frames in the memory to accurately predict the exact time.
In that sense, MATR preserves temporal information in memory to accurately localize actions and effectively stores information by using flag tokens.
A comparison with the OAD memory modules is shown in Table~\ref{table:oad_memory}.

\section{Proposed Method}

Consider an untrimmed video $\textbf{V}=\{\textbf{v}_i\}^T_{i=1}$ with $T$ frames and $M$ action instances $\Psi=\{(s_m,e_m,c_m)\}^M_{m=1}$, where each instance is represented by its start time $s_m$, end time $e_m$ and action class $c_m$. 
Unlike temporal action localization (TAL), which allows a model to use whole video sequence as input, online temporal action localization (On-TAL) forces a model to predict action instances using only input frames seen until the current timestamp. 
Note that action instances predicted at previous iterations cannot be modified or deleted afterwards.

\begin{figure}[t!]
\centering
\includegraphics[width=1.0\linewidth]{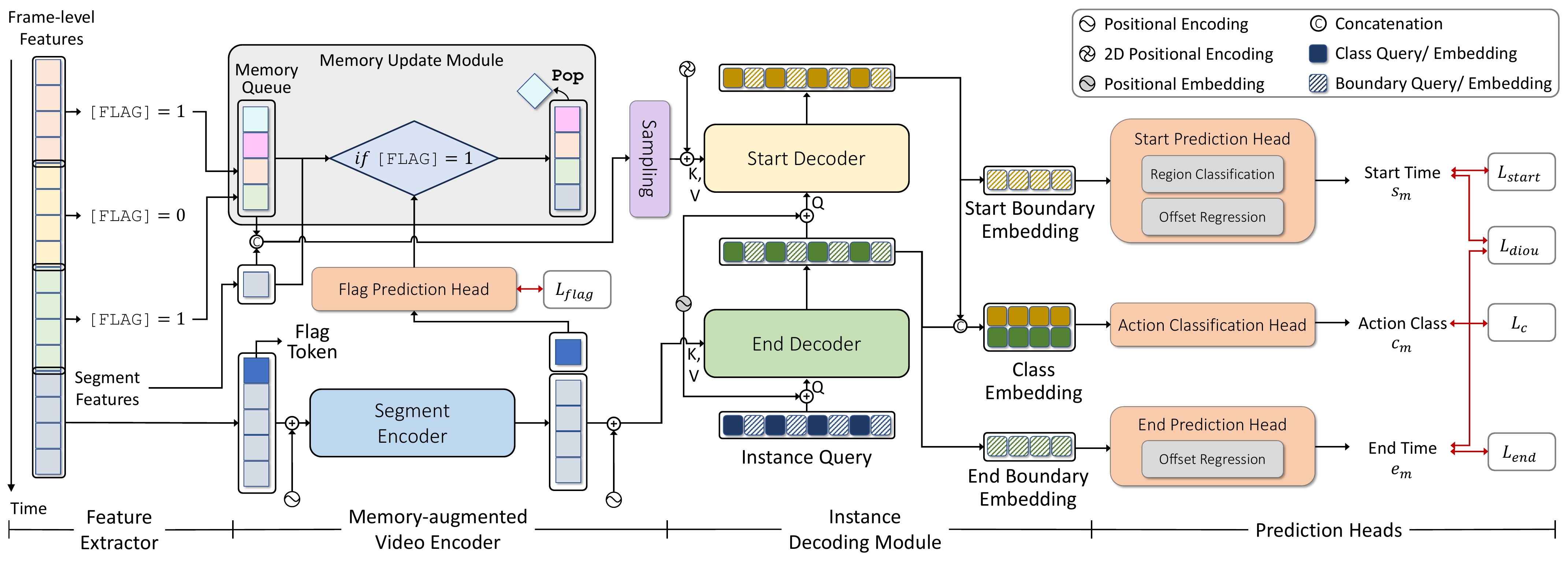}
\caption{Overall architecture of MATR. 
MATR consists of four parts: feature extractor, memory-augmented video encoder, instance decoding module, and prediction heads.
}

\label{fig:architecture}
\end{figure}

As shown in Fig.~\ref{fig:architecture}, MATR consists of four parts: feature extractor, memory-augmented video encoder, instance decoding module, and prediction heads. 
In line with previous research, the current segment of the input streaming video is given as a unit input, and its frame-level features, referred to as segment features, are extracted by a video backbone network and a linear projection layer (Sec.~\ref{sec:3.1}). 
The segment features are then fed to the memory-augmented video encoder, which encodes temporal context between frames in the current segment and stores the segment features into the memory (Sec.~\ref{sec:3.2}).
The instance decoding module localizes action instances via two Transformer decoders: the end decoder and the start decoder.
Specifically, the end decoder references the encoded segment features to locate the action end around the current time, and then the start decoder refers to the memory queue to find the action start based on the past information stored in the memory queue. 
Queries for each instance consist of a class query for action classification and a boundary query for action localization (Sec.~\ref{sec:3.3}).
The outputs of the instance decoding module are used as inputs to the prediction heads, which consist of end prediction head, start prediction head, and action classification head (Sec.~\ref{sec:3.4}).
The entire model is trained in an end-to-end manner (Sec.~\ref{sec:3.5}).

\subsection{Feature Extractor}
\label{sec:3.1}
Given a segment comprising $L_s$ consecutive video frames as input, a pretrained backbone~\cite{wang2016temporal,i3d} followed by a linear projection layer extracts the segment features $\textbf{X}_t=\{\textbf{x}_i\}^t_{t-L_s+1} \in \mathbb{R}^{L_s\times D}$ for each frame within the segment, following previous work~\cite{kang2021cag,kim2022sliding}. 
Inspired by Kim~\etal~\cite{kim2022sliding}, MATR employs a sliding window scheme, which predicts multiple action instances by moving frame by frame along the temporal axis. 
Note that the window size is the input segment size.

\subsection{Memory-Augmented Video Encoder}
\label{sec:3.2}
At each time step, the input segment features are fed to two modules of the memory-augmented video encoder, the segment encoder and the memory update module. 
The segment encoder encodes the temporal context across input segment features, while the memory update module selectively stores the segment features and updates the memory queue.

\noindent
\textbf{Segment Encoder.}
The segment encoder is a standard Transformer encoder composed of self-attention layers and a feed forward network (FFN).
The combination of segment features and a learnable flag token is fed into the encoder, and transformed to queries, keys, and values for self-attention.
Sinusoidal positional encoding $\textbf{S}_\text{pos} \in \mathbb{R}^{(L_s+1) \times D}$ is added to the queries and keys.
Among the output embeddings, the encoded flag token is fed into the memory update module and the encoded segment features that integrate temporal context across input frames are used as input to the end decoder.

\noindent
\textbf{Memory Update Module.}
In an online setting, where an input is a streaming video, efficiently storing information of past frames and accessing it effectively is crucial for detecting action instances, in particular those of long-term actions whose temporal extents are beyond the size of input segments. 
To this end, we adopt the memory queue for storing information of past input segments in a first-in-first-out (FIFO) manner. 
Additionally, we propose an efficient memory update method employing the flag token. 
When the input segment features pass through the segment encoder, the flag token is concatenated into the input segment features and then employed in the flag prediction head. 
It is trained to predict $[\texttt{FLAG}]$, which identifies whether an input segment is relevant to action instances, so that only relevant segments are stored in the memory.
$[\texttt{FLAG}]=1$ when there are overlapping frames between the input segment and action instances, and 0 otherwise.
During training, the ground-truth $[\texttt{FLAG}]$ is utilized.
During inference, $[\texttt{FLAG}] = 1$ if ${\text{sigmoid}(\hat{g}) > \theta}$ where $\hat{g}$ is the output logit of the flag prediction head and $\theta$ is a predefined threshold, and 0 otherwise.
The input segment feature is added to the memory queue when $[\texttt{FLAG}]=1$ and discards it when $[\texttt{FLAG}]=0$. 
If the memory queue is full, the oldest segment feature is purged.

\subsection{Instance Decoding Module} 
\label{sec:3.3}
\begin{figure}[t]
    \centering
    \includegraphics[width=\linewidth]{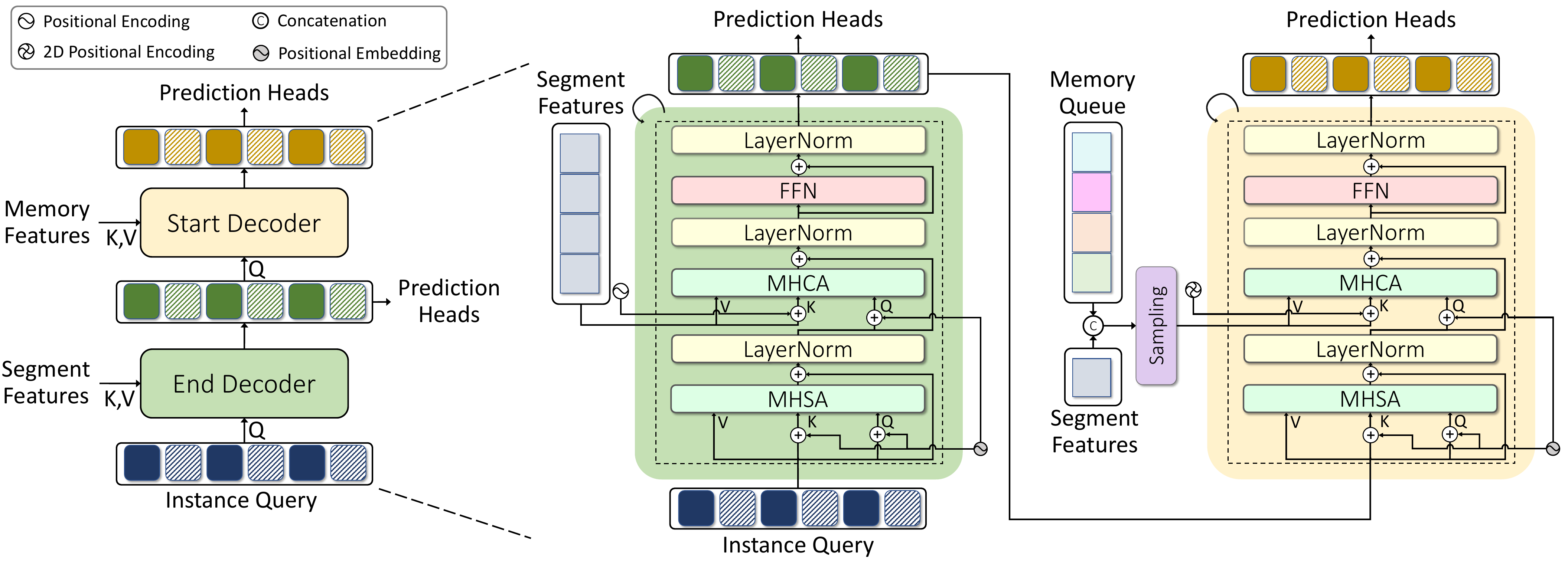}
    \caption{Detailed architecture of the instance decoding module.}
    \label{fig:decoder_architecture}
\end{figure}

The instance decoding module localizes and classifies action instances by leveraging the encoded segment features and the memory queue through the attention mechanism of Transformer.
Given the encoded segment features that have short-term temporal contexts across input segment and the memory queue that stores long-term contexts for actions that are potentially ongoing, a set of $2N$ instance queries $\textbf{Q}=\{\textbf{Q}_\text{class}; \textbf{Q}_\text{bound}\} \in \mathbb{R}^{2N\times D}$ is trained to generate $N$ action instances for each input segment. 
Half of these queries pertain to predicting the action class (class query; $\textbf{Q}_\text{class} \in \mathbb{R}^{N\times D}$), while the other half is used for predicting the start and end timestamps (boundary query; $\textbf{Q}_\text{bound} \in \mathbb{R}^{N\times D}$). 
The class and boundary query pairs for the same instance share the same positional embedding $\textbf{E}_\text{pos} \in \mathbb{R}^{N\times D}$ to accurately identify and differentiate between instances; $\tilde{\textbf{Q}}=\{\textbf{Q}_\text{class}+\textbf{E}_\text{pos};\textbf{Q}_\text{bound}+\textbf{E}_\text{pos}\}$.
Unlike the previous Transformer-based On-TAL method~\cite{kim2022sliding} that predicts start and end offset for each instance simultaneously, our method employs separate Transformer decoders for predicting start offset and end offset. 
As shown in \cref{fig:decoder_architecture}, the start decoder and end decoders share the same architecture but utilizes different information to predict action start and action end, respectively.

\noindent
\textbf{End Decoder.}
Given the encoded segment features, instance queries \textbf{Q} is trained to generate the output embeddings used for detecting action end near the current timestamp. 
To this end, we adopt a Transformer~\cite{carion2020end} composed of multi-head self-attention layers, multi-head cross-attention layers, and FFNs.
The detailed architecture is illustrated in Fig.~\ref{fig:decoder_architecture}.
The outputs of the end decoder, which consist of end boundary embeddings and class embeddings, are then used as inputs to the start decoder and the prediction heads.

\noindent
\textbf{Start Decoder.}
The start decoder takes the output embeddings from the end decoder and locates the corresponding action start by utilizing the memory queue.
First, the memory queue is concatenated with the current segment features.
The concatenated memory features are used as long-term context in the start decoder.
Then, 2D temporal positional encodings are added to the memory features, and the results are used as the \textit{key} and \textit{value} for the cross-attention layers of the start decoder.
The overall architecture of the start decoder is the same as the end decoder (Fig.~\ref{fig:decoder_architecture}), but it uses the memory features rather than the encoded segment features.
The output embeddings of the start decoder, start boundary embeddings and class embeddings, are used as inputs to the start prediction and the action classification head, respectively. 

Moreover, we introduce two techniques to efficiently and effectively utilize the memory queue. 
First, we conduct a 50\% uniform sampling to the memory features which enables efficient use of memory since adjacent frames contain similar information. 
Second, to expand the scope of positional encoding for streaming video, whose duration is unpredictable, we separate the temporal positional encoding into the two parts: relative segment position and relative frame position.
A relative segment position is the position of a segment within the memory relative to the current segment.
Meanwhile, a relative frame position is the position of a frame relative to the most recent frame within the same segment.

\subsection{Prediction Heads}
\label{sec:3.4}
To generate $N$ action instances $\{(\hat{s}_i,\hat{e}_i,\hat{c}_i)\}^N_{i=1}$ from the output of the instance decoding module, we use three prediction heads, which are end prediction head, start prediction head, and action classification head. 
Each of the prediction heads is composed of a 2-layer FFN.

\noindent
\textbf{End Prediction Head.}
The end prediction head estimates the offset between the end of the target action and the current timestamp.
The boundary embeddings from the end decoder are fed to the 2-layer FFN, which in turn estimates the end offset $\{\hat{u}_{i}\}_{i=1}^N \in \mathbb{R}^{N}$.

\noindent
\textbf{Start Prediction Head.}
The start prediction head estimates the offset between the start of the target action and the current timestamp. 
Unlike the end prediction, which finds the end offset around the input segment, the start prediction needs to perform offset regression over a relatively wider range. 
To narrow down the scope of the start offset regression, we divide the time horizon into $L_m+2$ regions: the region preceding the coverage of memory, $L_m$ regions covered by the memory, and the region corresponding to the current input segment. 
In that sense, the start prediction head consists of the region classification and the offset regression, each composed of a 2-layer FFN. 
They take the boundary embeddings from the start decoder as input to predict the start time. 
The region classification identifies the regions to which the start time is assigned, represented as $\{\hat{o}_i| \hat{o}_i = \text{argmax}(\hat{r}_i)\}^N_{i=1}  \in \mathbb{R}^N$. Here, $\{\hat{{r}_i}\}^N_{i=1} \in \mathbb{R}^{N \times (L_m+2)}$ are the output logits from the region classification head. Following this, the offset regression head is utilized to estimate the offset within the regions, denoted as $\{\hat{v}_i\}_{i=1}^N \in \mathbb{R}^{N \times (L_m+2)}$.
Note that the start offset $\hat{v}_i$ is calculated for all $L_m+2$ regions, and the offset from the identified region $\{\hat{v}_i(\hat{o}_i)\}^N_{i=1} \in \mathbb{R}^N$ is used in inference.
By combining start classification result $\hat{o}_i$ and start regression result $\hat{v}_i(\hat{o}_i)$, the start time is predicted. 
The effectiveness of the proposed start prediction head is validated in Table~\ref{table:prediction_heads}.

\noindent
\textbf{Action Classification Head.}
For action class prediction, we utilize the class embeddings from both the end decoder and the start decoder. 
These embeddings are concatenated and fed into the action classification head to derive class probabilities $\{\hat{p}_i\}_{i=1}^N \in \mathbb{R}^{N \times (C+1)}$, where $C$ is the number of action class.

\noindent
\textbf{Action Instance Prediction.}
Finally, our model makes $N$ action proposals $\{(\hat{s}_i,\hat{e}_i,\hat{c}_i)\}_{i=1}^N$ at time $t$ by
\begin{flalign} 
&\hat{s}_i = t-(\hat{o}_i+\hat{v}_{i}(\hat{o}_i)) \times L_s,\\
&\hat{e}_i = t+\hat{u}_{i} \times L_s, \\
&\hat{c}_i = \text{argmax}(\hat{p}_i),
\label{eqn:proposals}
\end{flalign}
where $L_s$ is the segment length.
Since the sliding window scheme generates $N$ action proposals at each timestamp, post-processing is crucial for performance improvement by removing redundant or overlapped action instances. 
Non-maximum suppression (NMS) is applied to the action proposals at each timestamp, subsequently removing those highly overlapped with the proposals generated from the past.
To prevent more reliable prediction made in the future from being removed, instances where the predicted end time is beyond the current timestamp, $t < \hat{e}_i$, are also removed.

\subsection{Training Objective}
\label{sec:3.5}
At timestamp $t$, the model is trained to detect action instances whose end times are in the range $[t-T_d+1, t+T_a]$, where $T_d$ and $T_a$ are hyperparameters. 
Then, hungarian algorithm matches the action proposals and the ground-truth action instances with the lowest matching cost. 
The matching cost of the ground-truth group $i$ and the proposal $\sigma(i)$ is given by:
\begin{equation}
C_{i, \sigma(i)} = \hat{p}_{\sigma(i)}(y_i)+\text{IoU}(b_i, \hat{b}_{\sigma(i)}),
\label{eq:matching_cost}
\end{equation}
where $\sigma$ is the permutation of $N$ action proposals, $\hat{p}_i$ is the class probabilities of the $i$-th proposal, $b_i = \{s_i, e_i\}$ is the ground-truth action boundary, and $\hat{b}_i = \{\hat{s}_i, \hat{e}_i\}$ is the predicted action boundary. 
We adopt the focal loss~\cite{lin2017focal} for the action classification, the cross entropy loss for the start region classification, the $\ell_1$ loss for both the start offset regression and the end offset regression. 
The action classification loss $L_\text{class}$, the start prediction loss $L_\text{start}$, and the end prediction loss $L_\text{end}$ are defined as follow: 
\begin{flalign} 
L_\text{class} &= \sum_{i=1}^N \text{FL}(\hat{p}_{i}, y_{i}),\\
L_\text{start} &= \sum_{i=1}^{N_\text{match}} \{\text{CE}(\text{Softmax}(\hat{r}_{i}), r_{i}) + |\hat{v}_{i}(\hat{o}_i) - v_{i}|\},\\
L_\text{end} &= \sum_{i=1}^{N_\text{match}} |\hat{u}_{i} - u_{i}|,
\label{eqn:losses}
\end{flalign}
where $y_i$, $r_i$, $v_i$, $u_i$ is the ground-truth action class, start region, start offset, and end offset, respectively.

To provide instance-level supervision and facilitate the connection between start and end time prediction, we employ the DIoU loss~\cite{zheng2020distance} following ActionFormer~\cite{zhang2022actionformer}:
\begin{equation}
    L_\text{diou} = \sum_{i=1}^{N_\text{match}} 1-\text{IoU}(\hat{b}_i, b_i)+ \frac{\rho^2(\hat{a}_i,a_i)}{d_i^2},
\end{equation}
where $\rho(\cdot, \cdot)$ is the Euclidean distance between two points, $\hat{a}_i$ and $a_i$ are the center of the proposal and ground-truth instance, and $d_i$ is the smallest enclosing 1-dim box length.

To train the memory update module in an end-to-end manner, we employ the flag loss for training the flag prediction head: 
\begin{equation}
    L_\text{flag} = \text{BCE}(\text{Sigmoid}(\hat{g}), g),
\end{equation}
where $\hat{g}$ and $g$ is the predicted logit of flag token and ground-truth $[\texttt{FLAG}]$ respectively. 

Our model is trained with five losses simultaneously in an end-to-end manner. 
The total loss is calculated as follows:
\begin{equation}
    L = L_\text{class} + L_\text{start} + L_\text{end} + L_\text{diou} + L_\text{flag}. 
\end{equation}
Note that all loss coefficients are set to $1$ and loss balancing is not required.

\section{Experiments}

\subsection{Experimental Setting}

\noindent
\textbf{Datasets.}
We evaluate our method and previous methods on two On-TAL benchmarks: THUMOS14~\cite{THUMOS14} and MUSES~\cite{MUSES}. 
THUMOS14 contains 200 videos for training and 213 videos for testing with 20 action classes, while MUSES has 2,587 videos for training and 1,110 videos for testing with 25 action classes.
MUSES consists of multi-shot action instances, which makes action localization more challenging.

\noindent
\textbf{Hyperparameters.}
We use a frozen two-stream TSN~\cite{wang2016temporal} pretrained on Kinetics~\cite{i3d} to extract RGB and flow features for THUMOS14, and an I3D~\cite{i3d} trained on Kinetics~\cite{i3d} utilizing only RGB features for MUSES, following previous On-TAL work~\cite{kang2021cag, tang2022simon, kim2022sliding}.
The segment size $L_s$ is set to 64 in THUMOS14 and 75 in MUSES.
The dimension of the segment feature is set to $D=1024$ for both datasets.
For the memory-augmented video encoder, we stack $3$ Transformer layers with $8$ attention heads for both datasets. 
The flag threshold $\theta$ is set to $0.5$. 
The memory size $L_m$ is set to 7 for THUMOS14, while 15 for MUSES. 
For the instance decoding module, we stack $5$ Transformer layers with $4$ attention heads for both datasets. 
The number of the class query and the boundary query $N=10$ for THUMOS14 and $N=6$ for MUSES. 

\noindent
\textbf{Training.}
During training, we use Adam optimizer~\cite{Adamsolver}, with the initial learning rate of $\expnum{1}{8}$ with CosineAnnealing scheduler~\cite{loshchilov2016sgdr}. 
The batch size is set to $64$ for THUMOS14, while $75$ for MUSES.
The focal loss coefficient is $\alpha=0.25$ and $\gamma=2$ for THUMOS14 and $\alpha=1$ and $\gamma=5$ for MUSES. 
All loss coefficients are set to 1.
More details are listed in the supplementary material ({Sec.}~D). 

\begin{table*}[t!]
\caption{Comparison with On-TAL methods on THUMOS14 and MUSES dataset. The results are reported in mAP measure (\%). '*' indicates that the result is validated using the On-TAL ground truth from classification annotation. }
\centering
\small
\scalebox{0.78}{
\begin{tabular}{lccccccccccccccc}
    \hline
    \multirow{2}{*}{Method} & \multicolumn{7}{c}{THUMOS14} && \multicolumn{7}{c}{MUSES} \\
    \cline{2-8} \cline{10-16}
                                & backbone    & 0.3   & 0.4   & 0.5   & 0.6   & 0.7   & Average &
                                & backbone    & 0.3   & 0.4   & 0.5   & 0.6   & 0.7   & Average \\
    \hline
    \multicolumn{15}{l}{\textbf{Offline TAL}}\\
    G-TAD~\cite{xu2020g}            &TSN~\cite{wang2016temporal} & 54.5  & 47.6  & 40.2  & 30.8  & 23.4  & 39.9  &
                                    & I3D~\cite{i3d} & 19.1  & 14.8  & 11.1  & 7.4   & 4.7   & 11.4  \\
    ContextLoc~\cite{zhu2021enriching}  & I3D & 68.3  & 63.8  & 54.3  & 41.8  & 26.2  & 50.9  &
                                        & -- & --    & --    & --    & --    & --    & --  \\
    P-GCN~\cite{zeng2019graph}      & I3D & 63.6    & 57.8    & 49.1    & --    & --    &  --   &
                                    & I3D & 19.9  & 17.1  & 13.1  & 9.7   & 5.4   & 13.0  \\                           
    MUSES~\cite{MUSES}              & I3D & 68.9  & 64.0  & 56.9  & 46.3  & 31.0  & 53.4  &
                                    & I3D & 25.9  & 22.6  & 18.9  & 15.0  & 10.6  & 18.6  \\
    ActionFormer~\cite{zhang2022actionformer}   & I3D & 82.1  & 77.8  & 71.0  & 59.4  & 43.9  & 66.8  &
                                                & -- & --    & --    & --    & --    & --    & --  \\
    TriDet~\cite{shi2023tridet}   & I3D & 83.6  & 80.1  & 72.9  & 62.4  & 47.4  & 69.3  &
                                                & -- & --    & --    & --    & --    & --    & --  \\
    \hline
    \multicolumn{15}{l}{\textbf{OAD-based Online TAL}}\\
    TeSTra~\cite{zhao2022real}      &TSN & 35.6  & 29.2  & 21.4  & 13.4  & 7.6   & 21.4  &
                                    & -- & --    & --    & --    & --    & --    & --   \\
    MAT~\cite{wang2023memory}       &I3D & 36.0  & 27.5  & 19.2  & 12.3  & 6.4   & 20.3  &
                                    & -- & --    & --    & --    & --    & --    & --   \\
    CAG-QIL~\cite{kang2021cag}      &TSN & 44.7  & 37.6  & 29.8  & 21.9  & 14.5  & 29.7  &
                                    &I3D & 8.5   & 6.5   & 4.2   & 2.8   & 1.9   & 4.8   \\
    SimOn~\cite{tang2022simon}      &TSN & 54.3  & 45.0  & 35.0  & 23.3  & 14.6  & 34.4  &
                                    & -- & --    & --    & --    & --    & --    & --    \\
    SimOn*~\cite{tang2022simon}     & TSN& 57.0  & 47.5  & 37.3  & 26.6  & 16.0  & 36.9  &
                                    & -- & --    & --    & --    & --    & --    & --    \\
    \hline
    \multicolumn{15}{l}{\textbf{Instance-level Online TAL}}\\
    OAT-Naive~\cite{kim2022sliding} &TSN & 57.6  & 50.6  & 43.0  & 30.0  & 15.7  & 39.4  &
                                    &I3D & 20.3  & 16.6  & 12.9  & 7.7   & 3.6   & 12.2  \\
    OAT-OSN~\cite{kim2022sliding}   &TSN & 63.0  & 56.7  & 47.1  & 36.3  & 20.0  & 44.6  &
                                    &I3D & 22.1  & 18.5  & 14.2  & 8.9   & 4.7   & 13.7  \\
    \rowcolor{Gray}
    MATR                            
    &TSN & \textbf{70.3}  & \textbf{62.7}  & \textbf{52.1}  & \textbf{38.6}  & \textbf{23.7}  & \textbf{49.5}  
    && I3D & \textbf{23.5} & \textbf{19.3} & \textbf{14.3}  & \textbf{9.4} & \textbf{5.7} & \textbf{14.4} \\
    \hline
\end{tabular}}
\label{table:thumos}
\end{table*}

\subsection{Comparison with the State of the Art}
We compare our method with previous online and offline TAL methods on THUMOS14 and MUSES dataset. 
As presented in Table~\ref{table:thumos}, our method outperforms all previous On-TAL methods in both benchmarks by a substantial margin: 4.9\%p of average mAP in THUMOS14 and 0.7\%p of average mAP in MUSES.
The results show that MATR utilizing the memory queue is more effective than previous OAD-based methods~\cite{zhao2022real,kang2021cag,tang2022simon,wang2023memory} and instance-level method OAT-OSN~\cite{kim2022sliding}.
We also evaluate two OAD methods utilizing memory~\cite{zhao2022real,wang2023memory} in On-TAL setting by grouping their frame-level predictions into action instances.
However, their performance is far lower than MATR, which verifies the effectiveness of the proposed memory queue of MATR.
We also compare our model with several offline TAL methods.  
Although there is still a performance gap between online and offline methods, our model achieves comparable performance to offline methods, and even outperforms previous work~\cite{xu2020g, zeng2019graph}.
The performance on the MUSES dataset is lower compared to THUMOS14 and the reason is twofold: 
(1) Action instances in MUSES are captured in multi-shots, making it challenging to detect action boundaries, and
(2) there are confusing action classes such as ``quarrel'' and ``fight,'' making action classification hard. 
These make MUSES challenging and limit the upper-bound of performance on it.

\subsection{Ablation Studies}
We perform ablation studies on THUMOS14 and MUSES dataset to show the effectiveness of the proposed modules. 

\begin{figure}[t]
\centering
\noindent
\begin{minipage}{.5\linewidth}
\vspace{-2.5mm}
\captionof{table}{Ablation of the proposed modules on THUMOS14 dataset. The results are reported in mAP measure (\%).}
\centering
\renewcommand{\arraystretch}{1.15}
\scalebox{0.74}{
\begin{tabular}{lcccccc}
    \hline
    Method                  & 0.3   & 0.4   & 0.5   & 0.6   & 0.7  & Average    \\
    \hline
    Ours                    & \textbf{70.3}  & \textbf{62.7}  & \textbf{52.1}  & 38.6  & \textbf{23.7}  & \textbf{49.5}      \\
    \hline
    \multicolumn{7}{l}{\textbf{(a) Memory-augmented video encoder}}\\
    w/o flag token          & 67.3  & 61.0  & 50.1  & 36.5  & 22.3  & 47.4      \\
    w/o segment encoder     & 65.8  & 59.3  & 49.9  & 36.3  & 21.8  & 46.6        \\
    \hline
    \multicolumn{7}{l}{\textbf{(b) Instance decoding module}}\\
    single decoder          & 65.1  & 56.4  & 45.0  & 30.5  & 16.7  & 42.7      \\
    w/o splitting query     & 66.2  & 61.4  & 50.9  & 38.0  & 23.2  & 47.9      \\
    w/o sampling            & 66.1  & 59.6  & 50.6  & \textbf{39.9} & 22.5  & 47.2      \\
    w/o pos embedding       & 67.2  & 60.5  & 49.8  & 36.3  & 22.7  & 47.3      \\
    \hline
    \multicolumn{7}{l}{\textbf{(c) Training objective}}\\
    w/o DIoU loss           & 62.2  & 53.4  & 43.2  & 31.0  & 17.4  & 41.4      \\
    \hline
\end{tabular}}
\label{table:ablation}
\end{minipage}
\begin{minipage}{.45\linewidth}
    \begin{minipage}{\linewidth}
        \captionof{table}{Analysis of the maximum length of the memory queue.}
        \centering
        \renewcommand{\arraystretch}{0.95}
        \scalebox{0.75}{
        \begin{tabular}{lcccccc}
            \hline
            Memory size         & 0.3   & 0.4   & 0.5   & 0.6   & 0.7  & Average \\
            \hline
            \multicolumn{7}{l}{\textbf{(a) THUMOS14 dataset}}\\
            w/o mem             & 65.8  & 58.9  & 47.6  & 36.9  & 20.8 & 46.0    \\
            1                   & 67.2  & 59.9  & 50.7  & 37.9  & 22.7 & 47.7    \\
            3                   & 67.3  & 61.9  & 51.9  & 37.9  & 22.6 & 48.3    \\
            7                   & \textbf{70.3} & \textbf{62.7} & 52.1  & 38.6  & 23.7 & \textbf{49.5}    \\
            11                  & 67.9  & 60.6  & 49.8  & 37.8  & 23.2 & 47.9    \\
            15                  & 66.9  & 60.1  & 50.7  & 38.4  & 24.1 & 48.0    \\
            19                  & 67.0  & 60.1  & \textbf{52.5}  & \textbf{40.2}  & \textbf{24.6} & 49.1    \\
            \hline
            \multicolumn{7}{l}{\textbf{(b) MUSES dataset}}\\
            w/o mem             & 22.3           & 17.6  & 13.1  & 9.1  & 4.6   & 13.3    \\
            1                   & 23.0           & 18.0  & 13.5  & 8.6  & 4.9   & 13.6    \\
            3                   & 23.1           & 18.3  & 13.9  & 8.9  & 5.1   & 13.9    \\
            7                   & \textbf{23.5}  & 18.1  & 13.5  & 8.9  & 4.9   & 13.8    \\
            11                  & 22.9           & 18.7  & 14.0  & 9.4  & 5.4   & 14.1    \\
            15                  & \textbf{23.5}  & \textbf{19.3}  & \textbf{14.3}  & 9.4   & \textbf{5.7}  & \textbf{14.4}    \\
            19                  & 22.7  & 18.9  & 13.8  & \textbf{9.6}   & 5.5  & 14.1    \\
            \hline
        \end{tabular}}
        \label{table:memory_size}
    \end{minipage}
\end{minipage}
\end{figure}

\noindent
\textbf{Ablation on the Proposed Modules.}
In Table~\ref{table:ablation}, we conduct an ablation study to verify the effectiveness of each component of our model. 
In Table~\ref{table:ablation})(a), focuses on the components of the memory-augmented video, both removing the flag token and the segment encoder led to a performance drop.
The results show the efficacy of selectively storing the past information and the importance of temporal context across the input segment. 
In Table~\ref{table:ablation}(b), we use a single decoder and predict both start and end at once, which results significant performance drop, from $49.5$ mAP to $42.7$ mAP.
The result demonstrates the contribution of our new action localization method, using two separate decoders for the start and the end prediction.
We also ablate new query design, sampling the memory for the start decoder, and positional embedding for instance queries. 
The results show that all design choices of the proposed components are important for action instance localization.
Lastly, training our model without DIoU loss shows a drastic performance drop, which demonstrates the importance of instance-level supervision for On-TAL (Table~\ref{table:ablation}(c)). 

\setlength{\tabcolsep}{3pt}
\begin{table}[t]
\centering
\caption{Comparison between different memory modules on the THUMOS14.}
\scalebox{0.8}{
\begin{tabular}{ccccccc}
    \hline
    Memory module  & Backbone & Segment size & Inference time  & fps & Memory parameters & Average mAP  \\
    \hline
    MAT~\cite{wang2023memory}          
            & TSN~\cite{wang2016temporal}  & 64           & 191.9ms          & 5.2      & 40.1M           & 46.9         \\
    E2E-LOAD~\cite{cao2023e2e}             
            & TSN  & 64           & 196.1ms          & 5.1      & 53.1M           & 47.9         \\
    MATR
            & TSN  & 64           & \textbf{167.1ms}           & \textbf{6.0}     & \textbf{24.0M}           & \textbf{49.5}         \\
    \hline
\end{tabular}}
\label{table:oad_memory} 
\end{table}

\noindent
\textbf{Impact of the Memory Queue Size.}
We conduct an ablation study on the different memory queue sizes.
As shown in Table~\ref{table:memory_size}, employing the memory queue enhances the performance compared to the without memory case, and generally the performance improves as the memory queue size increases. 
With an adequately large memory queue size covering the top $99\%$ of the instances in the training split, such as 7 for the THUMOS14 and 5 for the MUSES, the model shows robust performance and the performance is higher than the previous state-of-the-art model~\cite{kim2022sliding}.
The memory queue size of $7$ and $15$ shows the best on THUMOS14 and MUSES, respectively. 

\begin{figure}[t]
\centering
\noindent
\begin{minipage}{.4\linewidth}
    \centering
    \vspace{2.5mm}
    \includegraphics[width=\linewidth]{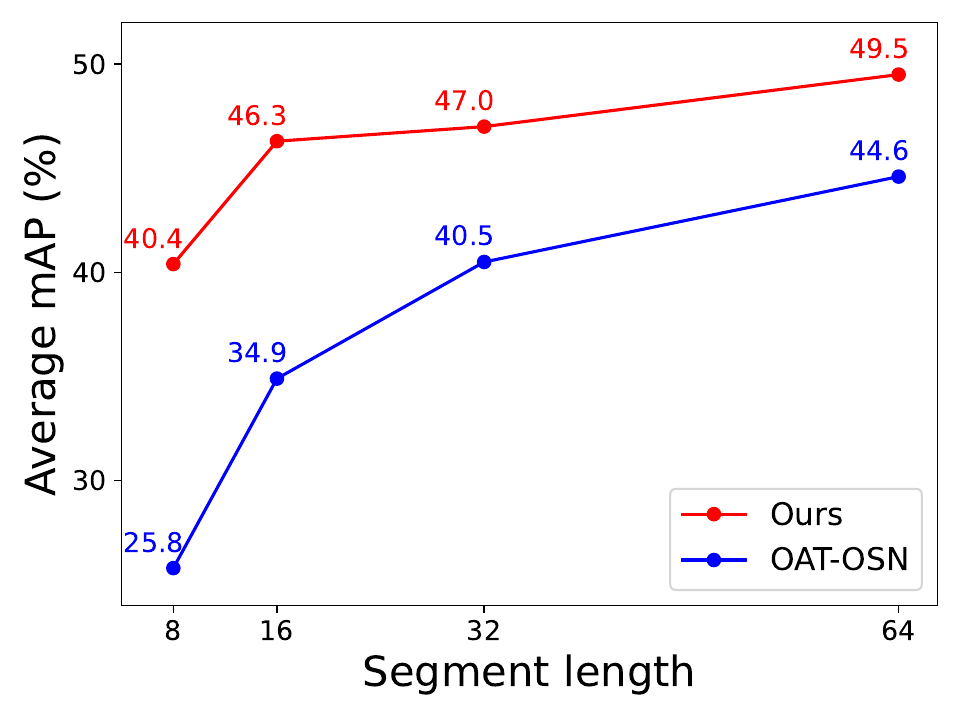}
    \captionof{figure}{Average mAP (\%) versus the segment size in THUMOS14. 
    Red line represents the result of our model and blue line represents the result of OAT-OSN.}
    \label{fig:seg_len}
\end{minipage}
\hfill
\begin{minipage}{.57\linewidth}
    \centering
    \begin{minipage}{\linewidth}
        \vspace{-2.5mm}
        \captionof{table}{Ablation on the start and the end prediction heads. Reg and Cls refer to the offset regression and the region classification respectively.}
        \centering
        \setlength{\tabcolsep}{1pt}
        \scalebox{0.8}{
        \begin{tabular}{cccccccc}
            \hline
            Start           & End          & 0.3   & 0.4   & 0.5   & 0.6   & 0.7  & Average \\
            \hline
            Reg             & Reg          & 64.2  & 59.0  & 50.0  & 37.8  & 22.7 & 46.7    \\
            \textbf{Reg + Cls}       & \textbf{Reg}          & \textbf{70.3}  & \textbf{62.7}  & \textbf{52.1}  & \textbf{38.6}  & \textbf{23.7} & \textbf{49.5}    \\
            Reg + Cls       & Reg + Cls    & 64.0  & 58.9  & 50.2  & 38.3  & 23.2 & 46.9    \\
            \hline
        \end{tabular}}
    \label{table:prediction_heads}
    \end{minipage}
    \begin{minipage}{\linewidth}
        \small
        \vspace{2mm}
        \captionof{table}{Ablation on the memory compression factor on the THUMOS14 dataset.}
        \centering
        \renewcommand{\arraystretch}{1.1}
        \setlength{\tabcolsep}{1.5pt}
        \scalebox{0.9}{
        \begin{tabular}{lcccccc}
            \hline
            Compress factor     & 0.3   & 0.4   & 0.5   & 0.6   & 0.7  & Average \\
            \hline
            None (Ours)         & \textbf{70.3}  & \textbf{62.7}  & 52.1  & 38.6  & 23.7 & \textbf{49.5}    \\
            2                   & 67.3  & 59.9  & 49.4  & 36.9  & 22.3 & 47.7    \\
            4                   & 68.1  & 61.5  & \textbf{52.7}  & \textbf{39.7}  & \textbf{24.5} & 49.3    \\
            8                   & 67.5  & 61.3  & 52.0  & 38.8  & 23.8 & 48.7    \\
            \hline
        \end{tabular}}
        \label{table:compress_factor}        
    \end{minipage}
\end{minipage}
\end{figure}

\setlength{\tabcolsep}{3pt}
\begin{table}[t]
\centering
\caption{Comparison of the inference time and \# parameters on the THUMOS14.}
\scalebox{0.8}{
\begin{tabular}{ccccccc}
    \hline
    Method  & Backbone & Segment size & Inference time   & fps      & \# Parameters    & Average mAP   \\
    \hline
    OAT-OSN~\cite{kim2022sliding}          
            & TSN~\cite{wang2016temporal}  & 64           & 163.7ms          & 6.1      & 128.7M           & 44.6         \\
    MATR             
            & TSN  & 64           & 167.1ms          & 6.0      & 192.8M           & 49.5         \\
    MATR             
            & TSN  & 16           & 53.8ms           & 18.6     & 192.8M           & 46.3         \\
    \hline
\end{tabular}}
\label{table:inference_time} 
\end{table}

\noindent
\textbf{Comparison of the proposed memory queue with OAD methods using memory.}
We compare our memory module with previous memory modules from OAD tasks~\cite{wang2023memory,cao2023e2e} by replacing the memory-augmented video encoder (Sec.~\ref{sec:3.2}) of our model. 
For a fair comparison, hidden dimension of the attention blocks for all methods is set to 1024. 
As shown in Table~\ref{table:oad_memory}, MATR outperforms the existing memory modules in terms of both mAP and space-time complexity. 
While existing memory modules use multiple attention blocks, our memory module consists of a segment encoder and a flag prediction head, making it superior in terms of parameters and inference speed.

\noindent
\textbf{Analysis of the Segment Size.}
As shown in Fig.~\ref{fig:seg_len}, our model is less sensitive to the input segment size, which should be carefully tuned across various datasets. 
While the performance of OAT-OSN drops significantly from $44.6$ mAP to $25.8$ mAP when the segment size is reduced from $64$ to $8$, the performance drop of ours is only $9.1\%$p.

\noindent
\textbf{Analysis of Start and End Prediction Heads.}
We investigate various offset prediction methods for the start prediction and the end prediction (Table~\ref{table:prediction_heads}). 
For the first method, start offset regression without region classification, the offset is normalized using the offset statistics from the training dataset. 
The second approach, which adopts region classification and offset regression for the start prediction, and only offset regression for the end prediction, exhibits the best performance. 
The reason for performance drop when using region classification for the end prediction is that frame-wise end classification is a complex problem, adversely affects the performance.

\noindent
\textbf{Analysis of Memory Compression Factor.}
We also conduct an experiment on efficiently storing input segment features in the memory by compressing them. 
We utilize a 1D convolution layer to compress the number of segment features. 
As shown in Table~\ref{table:compress_factor}, 4$\times$ compression yields competitive results with efficient memory usage. However, we decide not to use this method due to the efficient utilization of memory already achieved through the flag token and avoid to use additional hyperparameters that require optimization for each dataset.

\noindent
\textbf{Inference Time.}
As shown in Table~\ref{table:inference_time}, when using the same backbone and the segment size, the inference time of MATR is almost equal to that of OAT-OSN, yet its average mAP is significantly better. 
Moreover, by reducing the segment length to 16, MATR outperforms OAT-OSN in terms of both mAP and the inference time.
MATR uses more parameters due to the separation of the decoder into the start decoder and the end decoders.

\begin{figure}[t]
    \centering
    \includegraphics[width=1.0\linewidth]{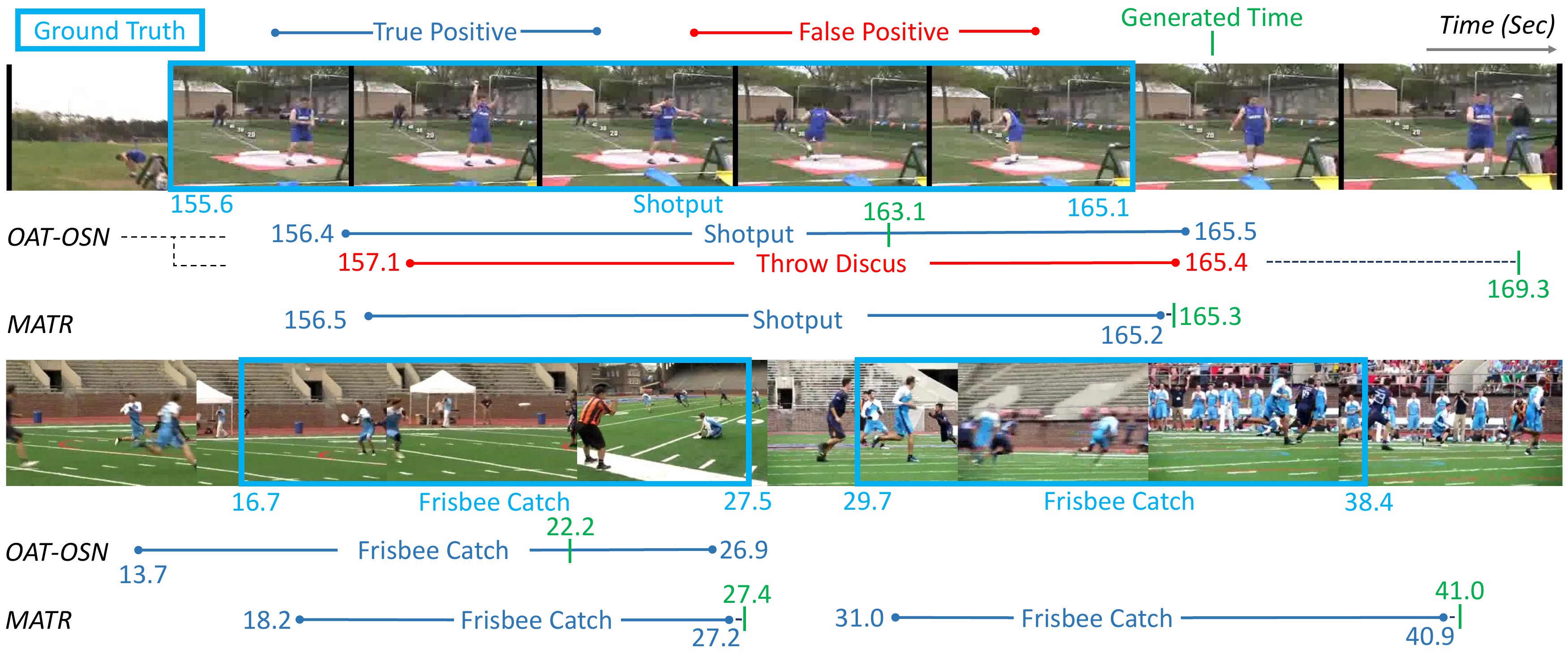}
    \caption{Qualitative results of MATR and OAT-OSN~\cite{kim2022sliding} on the THUMOS14. Generated time is the timestamp when the predicted instance is generated.}
    \label{fig:qualitative}
\end{figure}

\subsection{Qualitative Results}
Fig.~\ref{fig:qualitative} illustrates the predictions of our model and OAT-OSN~\cite{kim2022sliding}. 
The results show that MATR has the capability to efficiently detect action instances by identifying intervals immediately after the current action ends.
Additionally, the results show that our model is able to classify and localize action instances more precisely than OAT-OSN.

\section{Conclusion}
In this paper, we address online temporal action localization by introducing a new model, MATR, which leverages the memory queue to exploit long-term context. 
We also propose a new action instance localization method based on Transformer, which separately identifies action starts and action ends. 
As a result, MATR not only surpasses existing online TAL methods but also demonstrates comparable performance to offline TAL methods. 
The code base will be open to the public to promote future research on On-TAL.

\noindent \textbf{Limitations:} 
When the memory queue contains multiple action instances, there is a risk of action instances being matched with incorrect start points.
Additionally, when storing the input segment into the memory queue, MATR only considers whether the input segment is relevant to action instances but does not leverage the relationship with the past context stored in the memory.
To address these issues, investigating a method to utilize the information stored in the memory queue when storing the input segment features, could lead to more effective memory storage and utilization.

{\small
\noindent \textbf{Acknowledgement.} 
This work was supported by 
the NRF grant and 
the IITP grant             
funded by Ministry of Science and ICT, Korea
(RS-2019-II191906,          
 RS-2021-II212068,             
 RS-2022-II220290,              
 NRF-2021R1A2C3012728).         
}

%
%
\bibliographystyle{splncs04}
\bibliography{cvlab_video}

\clearpage
\beginsupplement

\noindent
\textbf{\Large Appendix}
\vspace{1em}

This supplementary material provides contents omitted in the main paper due to the page limit.
In Sec.~\ref{sec:A}, we provide further explanation regarding the 2D temporal positional encoding used in the start decoder of the instance decoding module. 
Sec.~\ref{sec:B} presents the memory queue update procedure.
In Sec.~\ref{sec:C}, we analyze AP scores for each class in the THUMOS14 and MUSES datasets.
Sec.~\ref{sec:D} describes additional experimental details including hyperparameter settings. 
Sec.~\ref{sec:E} presents in-depth analysis of our framework including the effects of the hyperparameters and the post-processing.
Sec.~\ref{sec:F} gives more qualitative results of our model on the THUMOS14 dataset.

\section{About 2D Temporal Positional Encoding}
\label{sec:A}

In this section, we provide a more detailed explanation regarding the implementation of the 2D temporal positional encoding, which is applied to the \textit{key} for the cross-attention layers of the start decoder in the instance decoding module. 
As explained in {Sec.}~3.3, the \textit{key} for the cross-attention layers consists of the memory queue concatenated with the current segment features. 
Therefore, the 2D temporal positional encoding should encompasses the relative position of segments within the memory queue and the relative position of the frames within the segment. 
Note that the relative position is defined with respect to the current frame and the position of the current segment features it belongs to.

To be specific, we explain a practical example of illustrating the application of the 2D temporal positional encoding.
Consider a scenario where the memory queue stores 4 segment features, and each segment is composed of 4 frames. 
If the indices of the segments stored in the memory queue are 3, 4, 9, and 10, with the current segment indexed as 13, then the indices for the memory features of the start decoder are 3, 4, 9, 10 and 13.
In this case, the relative segment positions from the current segment are 10, 9, 4, 3, and 0, respectively, while the relative frame positions within each segment are 3, 2, 1, and 0.
These two relative positions are encoded using sinusoidal positional encoding of dimension $D/2$ each, then concatenated to obtain the $D$-dimensional positional encoding.
By separating the relative segment positions from relative frame positions, the 2D temporal positional encoding becomes more efficient in representing positional encodings for a large number of frames. 

\section{Algorithm of Memory Update Module}
\label{sec:B}

\begin{figure}[t]
\noindent\scalebox{1}{
\begin{minipage}{\textwidth}
\begin{algorithm}[H]
\SetAlgoLined
    \PyCode{class UpdateMemQueue():} \\
    \Indp   
        \PyCode{self.mem\_queue = []} \PyComment{cached memory queue} \\
        \PyCode{self.max\_len} \PyComment{max memory len} \\
        \PyCode{} \\
        \PyCode{def forward(self, segment, flag):} \PyComment{update memory queue} \\
        \Indp
            \PyCode{if flag == True:} \\
            \Indp
                \PyCode{self.mem\_queue.append(segment)} \\
            \Indm
            \PyCode{if \text{len}(self.mem\_queue) $>$ self.max\_\text{len}:} \\
            \Indp
                \PyCode{self.mem\_queue.pop\_first()} \\
            \Indm
        \Indm        
    \Indm 
    \PyCode{} \\
    \PyCode{if training == True:} \PyComment{select flag type} \\
    \Indp
        \PyCode{flag = gt\_flag} \\ 
    \Indm
    \PyCode{else:} \\
    \Indp
        \PyCode{flag\_prob = flag\_token.sigmoid()} \\
        \PyCode{if flag\_prob $>$ flag\_threshold:}\\
            \Indp
            \PyCode{flag = True}\\
            \Indm 
        \PyCode{else:} \\
            \Indp
            \PyCode{flag = False}\\
    \Indm
    \PyCode{} \\
    \Indm
    \PyCode{if isNewVideo:} \PyComment{reset update module when a new video starts} \\
    \Indp
        \PyCode{update\_mem = UpdateMemQueue()} \\
    \Indm
    \PyCode{} \\    
    \PyCode{update\_mem(segment, flag)} \PyComment{update memory queue} \\
\caption{Pseudocode for memory update.}
\label{algo:supp_mem}
\end{algorithm}
\end{minipage}
}
\end{figure}

We present the details of memory update procedure in Algorithm~\ref{algo:supp_mem}.
The algorithm provides pseudocode of an ``UpdateMemQueue'' function and its usage in the memory queue update procedure. 
The input parameters of the function are \texttt{segment} and \texttt{flag} (line 23), where \texttt{segment} refers to the current segment features and \texttt{flag} is the Boolean flag to determine whether the current segment should be stored in the memory queue or not.
During training, the ground-truth flag is used (line 11-12).
During inference, the flag is true if the flag probability exceeds the flag threshold, and false otherwise (line 13-18). 
Note that the function is reinitialized at the start of each new video to prevent the mixing of the memory queues across different videos (line 20-21).

\section{Analysis of AP scores of MATR}
\label{sec:C}

\begin{figure}[p]
    \centering
    \includegraphics[width=0.9\linewidth]{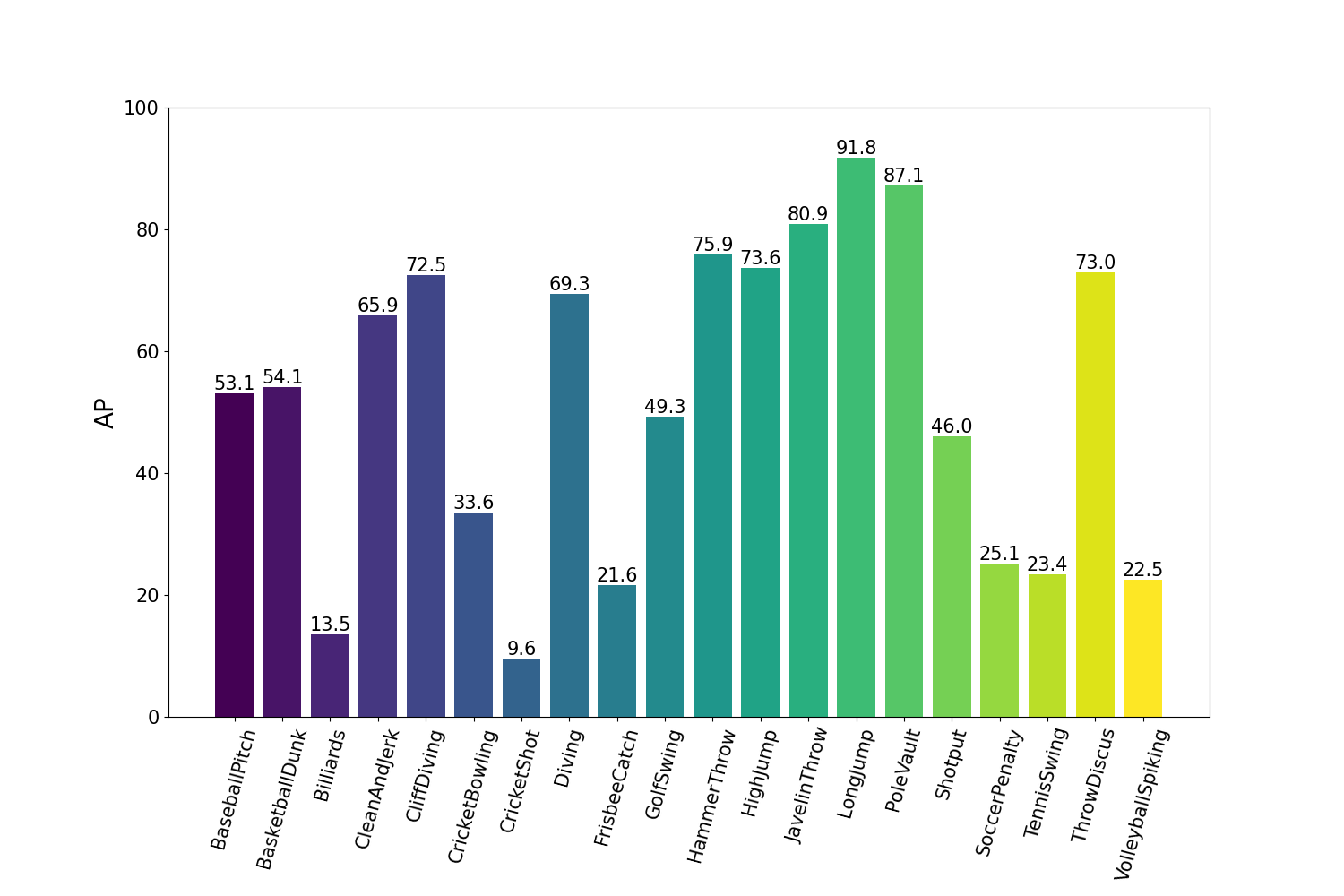}
    \caption{The Average Precision (AP) for each class at $\text{IoU threshold} = 0.5$ on the \\THUMOS14 dataset.}
    \label{fig:supp_thumosap}
    \centering
    \includegraphics[width=0.9\linewidth]{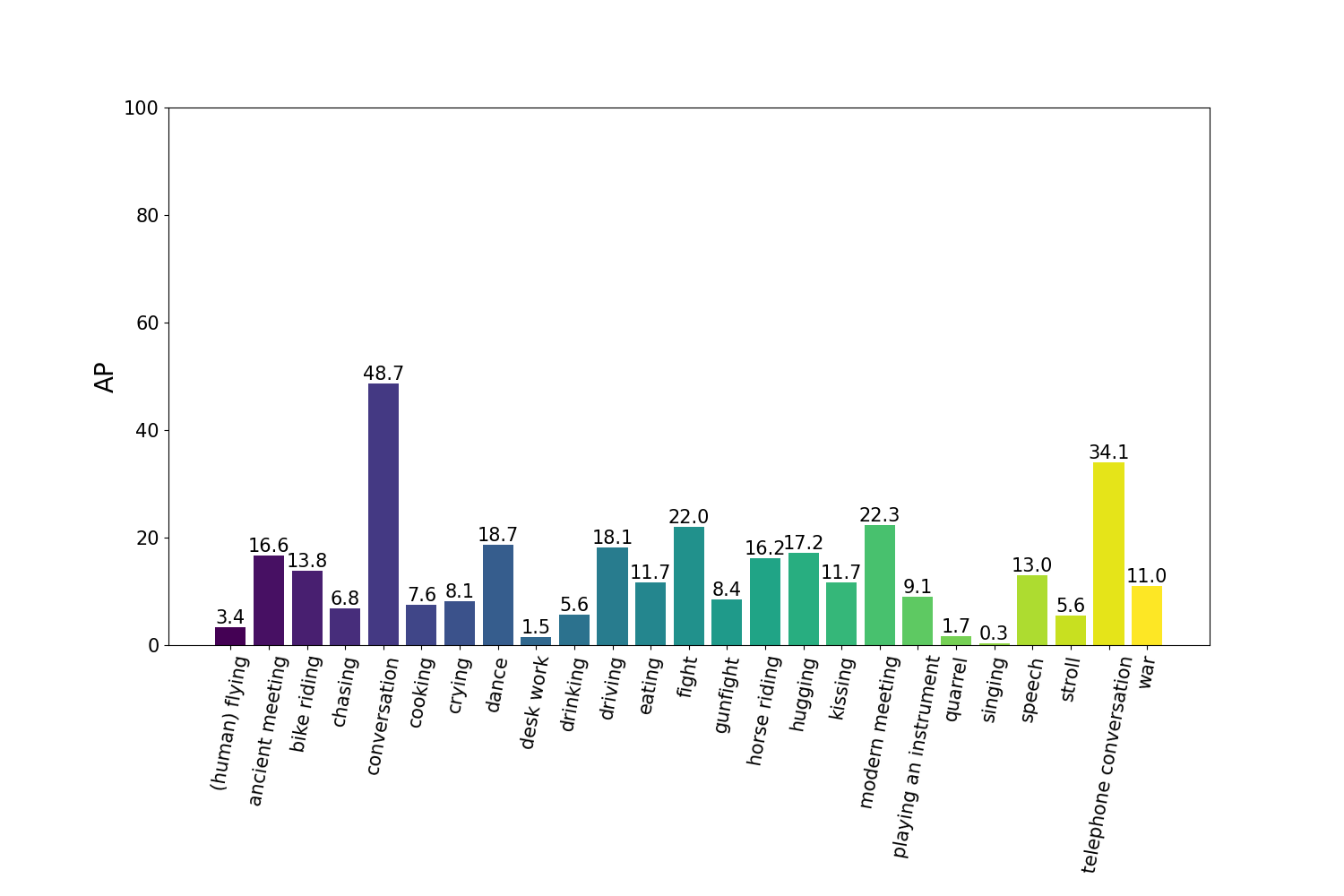}
    \caption{The Average Precision (AP) for each class at $\text{IoU threshold} = 0.5$ on the \\MUSES dataset.}
    \label{fig:supp_musesap}
\end{figure}

Fig.~\ref{fig:supp_thumosap} and Fig.~\ref{fig:supp_musesap} show the Average Precision (AP) scores of MATR on the THUMOS14~\cite{THUMOS14} and MUSES dataset~\cite{MUSES}. 
In the THUMOS14 dataset, action classes \textit{Long Jump}, \textit{Pole Vault}, and \textit{Javelin Throw} show the highest performance, while action classes \textit{Cricket Shot} and \textit{Billiards} show the lowest AP. 
In the MUSES dataset, the \textit{Conversation} class has the highest AP of $48.7$, while \textit{Singing} is measured with the lowest AP of $0.3$.
Both datasets encompass human-centric action classes, resulting in small differences between action classes and making it challenging to classify different action classes, requiring fine-grained classification. 
Specifically, the MUSES dataset has confusing action classes such as \textit{quarrel} and \textit{fight}, \textit{conversation} and \textit{telephone conversation}. 

\section{Experimental Details}
\label{sec:D}

\noindent
\textbf{Hyperparameters.}
We use a frozen two-stream TSN~\cite{wang2016temporal} pretrained on Kinetics~\cite{i3d} to extract RGB and flow features for THUMOS14, and an I3D~\cite{i3d} trained on Kinetics~\cite{i3d} utilizing only RGB features for MUSES, following previous On-TAL work~\cite{kang2021cag, tang2022simon, kim2022sliding}.
The segment size $L_s$ is set to 64 in THUMOS14 and 75 in MUSES.
The dimension of the segment feature is set to $D=1024$ for both datasets.
For the memory-augmented video encoder, we stack $3$ Transformer layers with $8$ attention heads for both datasets.
The flag threshold $\theta$ is set to $0.5$. 
For the instance decoding module, we stack $5$ Transformer layers with $4$ attention heads for both datasets. 
The number of the class query and the boundary query $N=10$ for THUMOS14 and $N=6$ for MUSES. 
We set the online NMS threshold used for post-processing to $0.3$.
The memory size $L_m$ is set to 7 for THUMOS14, while 15 for MUSES.
For the prediction heads, $T_d$ and $T_a$ are set to 16.

\noindent
\textbf{Training.}
We train our model with Adam optimizer~\cite{Adamsolver} with $\beta_1 = 0.9, \beta_2 = 0.999$, and $\epsilon = \expnum{1}{8}$ for 100 epochs on the THUMOS14 dataset and 30 epochs on the MUSES dataset.
We use the CosineAnnealing scheduler~\cite{loshchilov2016sgdr} with warmup restarts which have
the minimum learning rate of $\expnum{1}{8}$, the maximum learning rate of $\expnum{1}{5}$, $T_\text{cycle} = 10, T_\text{up} = 3$, and $\gamma = 0.9$, where $T_{cycle}$ is the number of epochs of one learning cycle, $T_{up}$ is the number of up-scaling epochs of one cycle, and $\gamma$ is the decreasing ratio of the $lr_{max}$ that decreases with each cycle.
The batch size is set to $64$ for THUMOS14, while $75$ for MUSES.
The focal loss coefficient is $\alpha=0.25$ and $\gamma=2$ for THUMOS14 and $\alpha=1$ and $\gamma=5$ for MUSES. 
All loss coefficients are set to 1.
\section{Additional Experiments}
\label{sec:E}

\setlength{\tabcolsep}{5pt}
\begin{table}[t]
\caption{Effects of the number of the instance queries $N$ of the instance decoding module on the THUMOS14 dataset.}
\label{table:num_query}
\centering
\begin{tabular}{lcccccc}
    \hline
    $N$             & 0.3   & 0.4   & 0.5   & 0.6   & 0.7  & Average \\
    \hline
    10 (Ours)           & \textbf{70.3}  & \textbf{62.7}  & \textbf{52.1}  & \textbf{38.6}  & \textbf{23.7}  & \textbf{49.5}    \\
    20                  & 66.6  & 61.0  & 51.4  & 37.3  & 23.1  & 47.9    \\
    30                  & 64.8  & 59.4  & 48.8  & 35.9  & 21.1  & 46.0    \\
    \hline
\end{tabular}
\end{table}

\noindent
\textbf{Analysis of the number of instance queries.}
In Table~\ref{table:num_query}, we investigate the effects of the number of instance queries $N$ in the instance decoding module. 
The performance tends to decrease with the number of instance queries in general, the optimal result is gained when $N=10$. 
Note that 10 is the smallest possible $N$, since $N$ should be greater than or equal to the maximum number of action instances that appear within the specified range $[t-T_d+1, t+T_a]$, where $T_d$ refers to the size of detection regions before the current timestamp $t$ and $T_a$ refers to the size of anticipation regions after the current timestamp $t$.

\setlength{\tabcolsep}{3.5pt}
\begin{table}[t]
\caption{Performance analysis according to the NMS threshold on the THUMOS14 dataset.}
\label{table:nms_threshold}
\centering
\begin{tabular}{lcccccc}
    \hline
    NMS threshold       & 0.3   & 0.4   & 0.5   & 0.6   & 0.7  & Average \\
    \hline
    0.1                 & 68.7  & 61.1  & 50.4  & 37.3  & 23.7  & 48.2    \\
    0.2                 & 69.8  & 61.9  & 51.3  & 37.9  & 23.4  & 48.9    \\
    0.3 (Ours)          & \textbf{70.3}  & \textbf{62.7}  & 52.1  & 38.6  & 23.7  & \textbf{49.5}    \\
    0.4                 & 69.6  & 62.3  & \textbf{52.3}  & 38.5  & 23.4  & 49.2    \\
    0.5                 & 67.8  & 61.0  & 51.6  & 38.6  & 23.7  & 48.5    \\
    0.6                 & 66.2  & 60.0  & 51.0  & \textbf{38.9}  & \textbf{24.6}  & 48.1    \\
    0.7                 & 62.6  & 57.0  & 49.0  & 37.9  & 24.4  & 46.2    \\
    \hline
\end{tabular}
\end{table}

\setlength{\tabcolsep}{3pt}
\begin{table}[t]
\caption{Analysis of the detection region size $T_d$ and the anticipation region size $T_a$ on the THUMOS14 dataset.}
\label{table:window_size}
\centering
\begin{tabular}{lcccccccc}
    \hline
    Method      & $T_d$   & $T_a$     & 0.3   & 0.4   & 0.5   & 0.6   & 0.7  & Average \\
    \hline
    A           & 8       & 8         & 67.5  & 62.1  & \textbf{52.1}  & 38.5  & 23.1  & 48.7    \\
    B           & 16      & 0         & 60.5  & 54.1  & 41.5  & 29.5  & 15.4  & 40.2    \\
    C           & 16      & 8         & 67.4  & 61.8  & 51.0  & 37.8  & 24.1  & 48.4    \\
    D (Ours)    & 16      & 16        & \textbf{70.3}  & \textbf{62.7}  & \textbf{52.1}  & \textbf{38.6}  & 23.7  & \textbf{49.5}    \\
    E           & 24      & 8         & 67.3  & 60.9  & 51.7  & 39.1  & \textbf{24.6}  & 48.7    \\
    F           & 32      & 0         & 63.2  & 54.9  & 43.1  & 28.9  & 17.4  & 41.5    \\
    \hline
\end{tabular}
\end{table}

\noindent
\textbf{Analysis of the post-processing.}
Table~\ref{table:nms_threshold} summarizes the effects of the NMS post-processing. 
As explained in {Sec.}~3.4 of the main paper, we adopt two NMS post-processing for MATR. 
One is NMS among the action proposals generated at the current timestamp, and the other is NMS among the highly overlapped action proposals generated from the past. 
The NMS threshold in Table~\ref{table:nms_threshold} represents the tIoU threshold used by the both NMS post-processing. 
Action instances surpassing the tIoU threshold are considered as the same instances and removed.
Our model performs robustly across various NMS thresholds, the optimal result is obtained when the threshold is $0.3$.

\noindent
\textbf{Analysis of the $T_d$ and $T_a$.}
As presented in Table~\ref{table:window_size}, we show the effects of the hyperparameters $T_d$ and $T_a$, where $T_d$ refers to the size of detection regions before the current timestamp $t$ and $T_a$ refers to the size of anticipation regions after the current timestamp $t$. 
As described in the {Sec.}~3.5 of the main paper, MATR is trained to predict action instances with action end time is in the range $[t-T_d+1, t+T_a]$.
Comparing A with B and D, E with F, even with the same total range length $T_d + T_a$, predicting action instances that end after the current timestamp ($e_m > t$) is very effective. 
In the comparison between B, C, and D with the same $T_d$, mAP increases as $T_a$ increases. 
The best performance is achieved when both $T_d$ and $T_a$ is set to 16. 
Therefore, we adopt the hyperparameter setting D for the final model. 

\section{More Qualitative Results}
\label{sec:F}

\begin{figure}[t]
    \centering
    \includegraphics[width=1\linewidth]{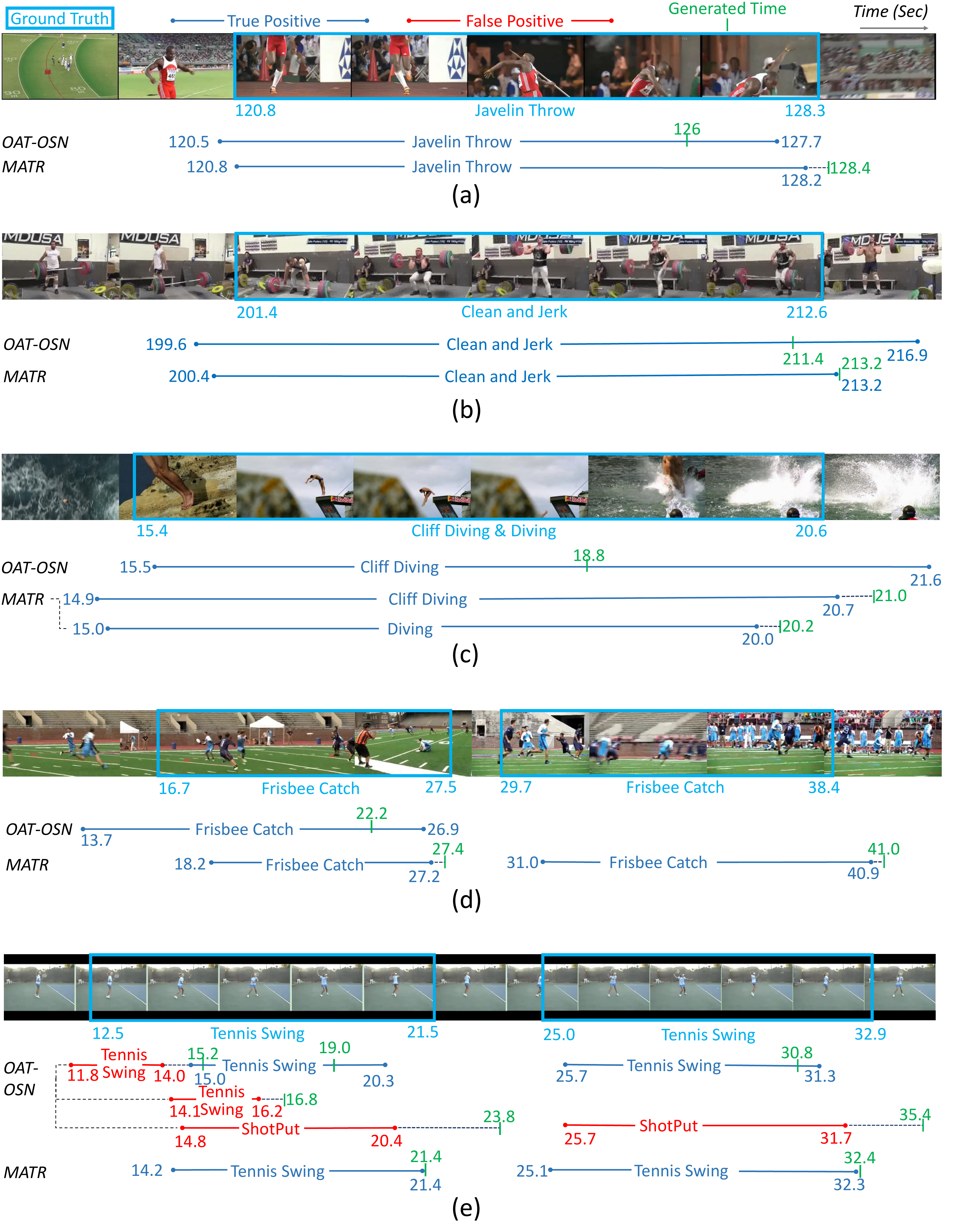}
    \caption{More qualitative results on the THUMOS14 dataset. 
    Generated time is the timestamp when the predicted instance is generated.}
    \label{fig:supp_qualitative}
\end{figure}

Fig.~\ref{fig:supp_qualitative} visualizes the predictions of MATR and OAT-OSN~\cite{kim2022sliding}. 
The results demonstrate that MATR is able to localize action instances better than existing On-TAL methods. 
In Fig.~\ref{fig:supp_qualitative}~(a) and (b), MATR accurately predicting the action end time by avoiding uncertain future predictions since MATR removes action instances where the predicted end time is beyond the current timestamp.
In Fig.~\ref{fig:supp_qualitative}~(c), MATR predicts instances of different classes having similar time intervals, such as \textit{Cliff Diving} and \textit{Diving}.
In Fig.~\ref{fig:supp_qualitative}~(d), MATR detects the second \textit{Frisbee Catch} action, which is challenging to detect solely based on the visual features from the current segment, by leveraging information regarding the previous \textit{Frisbee Catch} action instances stored in the memory queue as long-term context. 
In Fig.~\ref{fig:supp_qualitative}~(e), MATR operates robustly even on instances that are prone to being segmented into multiple parts.

\begin{figure*}[t]
    \centering
    \includegraphics[width=1\linewidth]{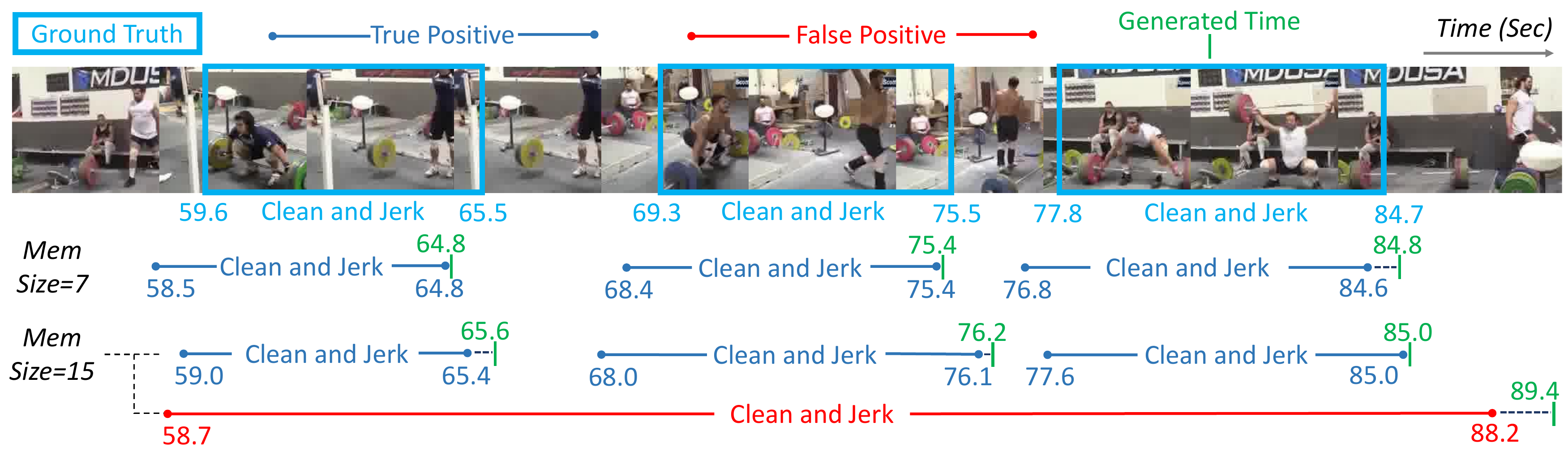}
    \caption{Qualitative results of MATR with different memory sizes on the THUMOS14.}
    \label{fig:qual_memory}
\end{figure*}

\noindent
\textbf{Limitations}
As shown in Fig.~\ref{fig:qual_memory}, when the memory queue contains multiple action instances, there is a risk of action instances being matched with incorrect start points.
When the memory sizes are both 7 and 15, the model predicts accurate actions and intervals for all three ground-truth instances.
However, with a memory size of 15, the memory queue contain more start timestamps with similar appearances, leading to additional instance being incorrectly matched with incorrect start timestamp.

\end{document}